\documentclass[10pt,twocolumn,letterpaper]{article}

\usepackage{wacv}
\usepackage{times}
\usepackage{epsfig}
\usepackage{graphicx}
\usepackage{amsmath}
\usepackage{amssymb}
\usepackage{multirow}
\usepackage{color}
\usepackage{bm}
\usepackage{graphicx}
\usepackage{subfigure}
\usepackage{caption}

\wacvfinalcopy 

\def\wacvPaperID{345} 

\ifwacvfinal\fi
\begin{document}
\title{An Enhanced Deep Feature Representation for Person Re-identification\thanks{Citation for this paper: Shangxuan Wu, Ying-Cong Chen, Xiang Li, An-Cong Wu, Jin-Jie You, and Wei-Shi Zheng. An Enhanced Deep Feature Representation for Person Re-identification. In IEEE WACV, 2016.}}


\author{Shangxuan Wu \hspace{0.5cm} Ying-Cong Chen\hspace{0.5cm} Xiang Li\hspace{0.5cm} An-Cong Wu\\
Jin-Jie You \hspace{1cm} Wei-Shi Zheng\thanks{Corresponding author}\\
Intelligence Science and System Lab, Sun Yat-sen University, China\\
Guangdong Provincial Key Laboratory of Computational Science, China\\
{\tt\small alanwsx@outlook.com, chyingc@mail2.sysu.edu.cn, lixiang651@gmail.com}\\
{\tt\small wuancong@mail2.sysu.edu.cn, youjinjie9@gmail.com, wszheng@ieee.org}
}
\maketitle
\ifwacvfinal\thispagestyle{empty}\fi

\setlength{\abovecaptionskip}{0pt}
\setlength{\belowcaptionskip}{0pt}

\begin{abstract}
   Feature representation and metric learning are two critical components in person re-identification models.
   In this paper, we focus on the feature representation and claim that hand-crafted histogram features can be complementary to Convolutional Neural Network (CNN) features.
   We propose a novel feature extraction model called Feature Fusion Net (FFN) for pedestrian image representation.
   In FFN, back propagation makes CNN features constrained by the hand-crafted features.
    Utilizing color histogram features (RGB, HSV, YCbCr, Lab and YIQ) and texture features
    (multi-scale and multi-orientation Gabor features), we get a new deep feature representation that is more discriminative and compact.
    Experiments on three challenging datasets (VIPeR, CUHK01, PRID450s) validates the effectiveness of our proposal.

\end{abstract}
\section{Introduction}
Person re-identification aims at matching people from d\nolinebreak[4]ifferent views under surveillance cameras, which has been studied extensively in the past five years.
To address the re-identification problem, existing methods exploit either cross-view invariant features
~\cite{gray2008viewpoint,farenzena2010person,zhang2011gabor,ma2012local,kviatkovsky2013color,zheng2013reidentification,koestinger2012large,ma2014covariance,liao2015person}
 or cross-view robust metrics~\cite{davis2007information,dikmen2011pedestrian,koestinger2012large,li2013learning,zheng2013reidentification,pedagadi2013local,chen2015mirror,zhao2013person,zhengtowards,wang2015cross}.

Recently, Convolutional Neural Network (CNN) have been adopted in person re-identification,~\eg~\cite{li2014deepreid,ahmed2015improved,yi2014deep,hu2015deep}.
Deep Learning provides a powerful and adaptive ap\nolinebreak[4]proach to handle computer vision problems without excessive handcraft on image features.
The back propagation algorithm dynamically adjusts the parameters in CNN, which unifies both feature extraction and pairwise comparison process in a single network.

\begin{figure}[htbp]
\centering {
\subfigure[VIPeR]
{
    \includegraphics[width=0.3\linewidth,height=0.55\linewidth]{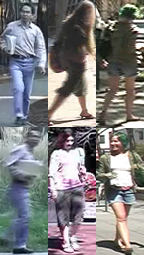}
}
\subfigure[CUHK01]
{
    \includegraphics[width=0.3\linewidth,height=0.55\linewidth]{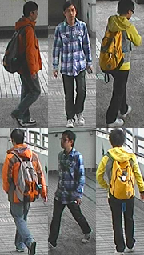}
}
\subfigure[PRID450s]
{
    \includegraphics[width=0.3\linewidth,height=0.55\linewidth]{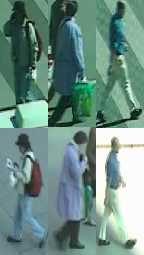}
}
}
\caption{Sample images from VIPeR, CUHK01 and PRID450s datasets. Images on the same column represent the same person.}
\label{fig: datasets_demo}
\end{figure}


However, in real-world person re-identification, a person's ap\nolinebreak[4]pearance often undergoes large variations across non-overlapping camera views, due to significant changes in view angle, lighting, background clutter and occlusion (see Fig.~\ref{fig: datasets_demo}).
Hand-crafted concatenation of different appearance features,~\eg RGB, HSV colorspaces and LBP descriptor, which are designed to overcome cross-view appearance variations in re-identification tasks, sometimes would be more distinctive and reliable.

In order to effectively combine hand-crafted features and deeply learned features,
we investigate the combination and complementary of a multi-colorspace hand-crafted features (ELF16) and deep features extracted from CNN.
A deep feature fusion Network (FFN)  is proposed in order to use hand-crafted features to regularize the CNN process so as to make the convolution neural networks extract features complementary to hand-crafted features. After extracting features by our FFN, traditional metric learning methods can be applied to boost the performance.
Experimental results on three challenging person re-identification datasets (VIPeR, CUHK01, PRID450s) demonstrate the e\nolinebreak[4]ffectiveness of our new features.
A significant improvemen\nolinebreak[4]t of Rank-1 matching rate is achieved as compared to state-of-the-art methods (8.09\%, 7.98\% and 11.2\%) on the three datasets.
In a word, we show that hand-crafted features could improve the extraction process of CNN features in FFN, achieving a more robust image representation.

\section{Related Works}

\noindent {\bf Hand-crafted Features.}
Color and texture are two of the most useful characteristics in image representation.
For example, HSV and LAB color histograms are used to measure the color information in the image.
LBP histogram~\cite{ojala2002multiresolution} and Gabor filter describe the textures of images.
Recent papers use a combination of different features to produce more effective features~\cite{zhang2011gabor,gray2008viewpoint,farenzena2010person,gray2008viewpoint,zheng2011person,zheng2013reidentification,ma2014covariance}.

Recently, features specifically designed for person re-identification significantly boost the matching rate.
 Local descriptors encoded by Fisher Vectors (LDFV)~\cite{ma2012local} build descriptors on Fisher Vector.
 Color invariants (ColorInv)~\cite{kviatkovsky2013color} use color distributions as the sole cue for good recognition performance.
Symmetry-driven accumulation of local features (SDALF)~\cite{farenzena2010person} proves that symmetry structure of segments can improve the performance significantly, and an accumulative method of features provides robustness to image distortions.
Local maximal occurrence features (LOMO)~\cite{liao2015person} analyzes the horizontal occurrence of local features and maximizes the occurrence to stably represent re-identification images.
%

\vspace{0.2cm}

\noindent {\bf Deep Learning.}
Convolutional Neural Network has been widely used in many computer vision problems, but only a few papers concern deep learning on person re-identification.

Li~\etal first proposed deep filter pairing nerual network (FPNN)~\cite{li2014deepreid} which used patch-matching layer and maxout pooling layer to handle pose and viewpoint variant.
FPNN was also the first work to employ deep learning on person re-identification problems.
Ahmed~\etal improved deep learning architecture by specifically designing cross-input neighbourhood difference layer~\cite{ahmed2015improved}.
Later, the deep metric learning in~\cite{yi2014deep} used ``siamese'' deep neural structure and a cosine layer to deal with big variations of person images.
Hu~\etal proposed deep transfer metric learning (DTML)~\cite{hu2015deep}, which transfers cross-domain visual knowledge into target datasets.

These deep methods combine feature extraction and image-pair classification into a single CNN network.
Pairwise comparison and symmetry structures are widely used among them, which could be inheritances of traditional metric learning methods~\cite{gray2008viewpoint,farenzena2010person,zhang2011gabor,ma2012local,kviatkovsky2013color,zheng2013reidentification,koestinger2012large,ma2014covariance,liao2015person,zhengtowards,wang2015cross}. Since pairwise comparison is form to learn the deep neural network, it is demanded to form quite a lot of pairs for each probe image and perform deep convolution on these pairs. Compared to these works, our FFN is not based on pairwise input but directly extracts deep features on a single image, so that our deep architecture can be followed by any conventional classifiers, while existing deep learning works cannnot.
%
%
%
%
\section{Methodology}

\subsection{Network Architecture}

\begin{figure*}
\begin{center}
\includegraphics[width=1.0\linewidth]
{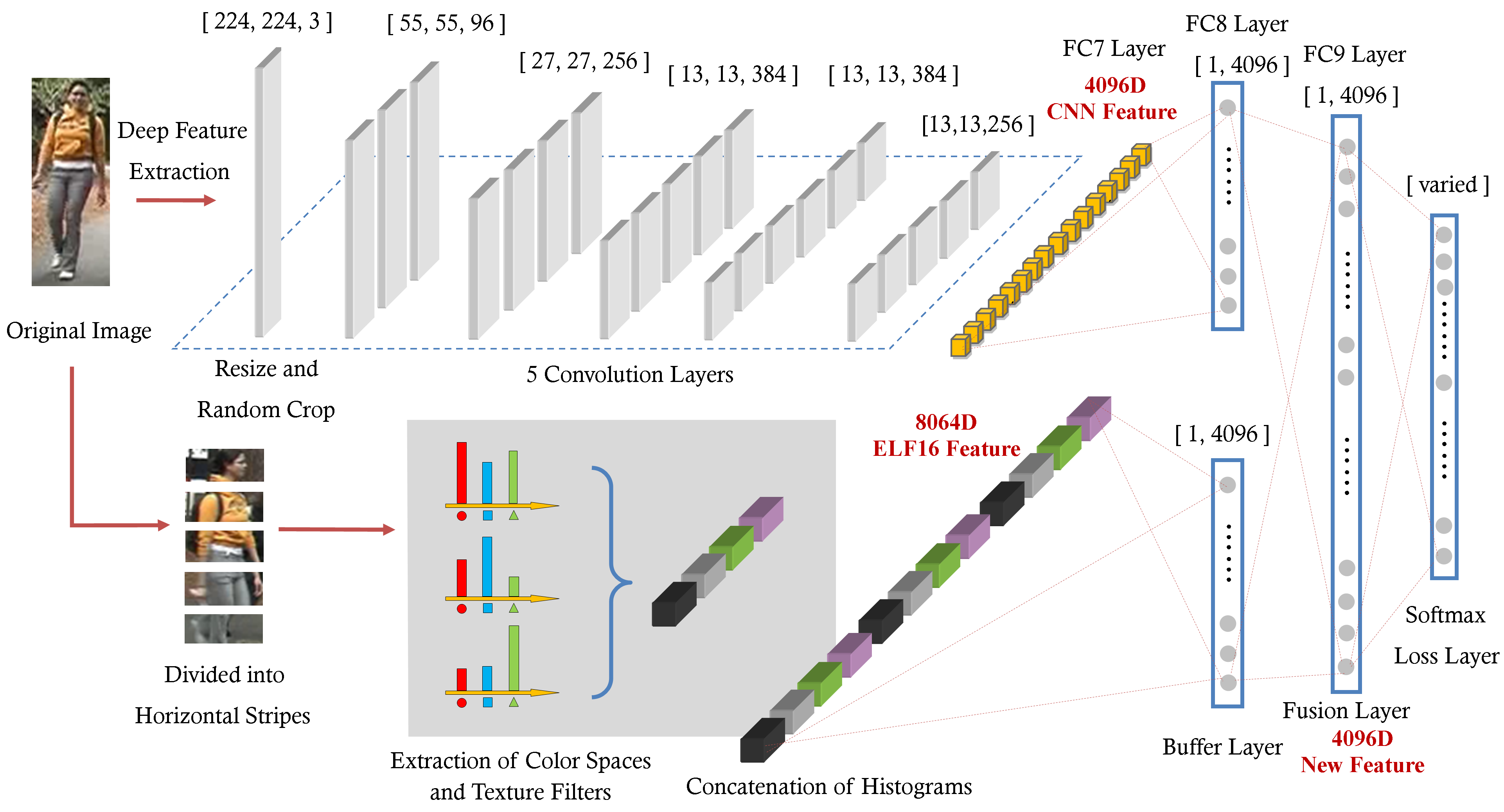}
\end{center}
   \caption{Fusion Feature Net (FFN) for ELF16 features and CNN features.}
\label{fig:fusion}
\end{figure*}

We use our modification of convolutional neural network (Feature Fu\nolinebreak[4]sion Net, FFN) to learn new features.
The network architecture is shown in Fig.~\ref{fig:fusion}.
Our Feature Fu\nolinebreak[4]sion Network consists of two parts. The first part deals with traditional convolution, pooling and activation neurons for input images; the second part processes additional hand-crafted feature representations of the same image.
These two sub-networks are finally linked together to produce a full-fledged image description, so the second part will regularize the first part during learning.
Finally, our new feature (4096D vector) is extracted from the last Full Convolution Layer (Fusion Layer) of FFN.

\subsection{CNN Features}
The upper part of Fig.~\ref{fig:fusion} describes a traditional process of convolution and pooling.
Every convolution layer is followed by a pooling layer and a local response normalizaion (LRN) layer~\cite{krizhevsky2012imagenet}, except for the $3^{rd}$ layer.
Finally, the output of the $5^{th}$ pooling layer is a 4096D vector, which we regarded as CNN Features.

Most re-identification models regard CNN as a whole binary classifier with direct image input like DeepReID~\cite{li2014deepreid} and Ahmed's Improved Deep Re-id Model~\cite{ahmed2015improved}.
However, the work in ~\cite{ding2015deep} inspires us to come up with strong reason for taking the convolution layer as a feature extractor.
One major characteristic of Re-identification images are whole-body images under different camera views.
Most of the body parts could be found in all the camera views, but suffer from serious malposition, distortion and misalignment.
The convolution in CNN allows part displacement and visual changes to be alleviated in higher-level convolution layers.
Multiple convolution kernels provide different descriptions for pedestrian images.
In addition, pooling and LRN layers provide nonlinear expression of corresponding description, which significantly reduces the overfitting problem.
These layers contribute to a stable Convolution Neural Network that could be applied to new datasets (See Section~\ref{section:training strategy} for detailed training process).

\subsection{Hand-crafted Features}
The lower part of Fig.\ref{fig:fusion} extracts conventional hand-crafted features widely used in person re-identification.
In this work, we employ the Ensemble of Local Features (ELF)~\cite{gray2008viewpoint} and is improved in ~\cite{zheng2011person,zheng2013reidentification}.
It extracts RGB, HSV and YCbCr histograms of 6 horizontal stripes of input image.
Also, 8 Garbor filters and 13 Schmid filters are applied to get corresponding texture information.

We modify ELF feature by improving the color space and stripe division~\cite{chen2015mirror}.
Input image is equally partitioned into 16 horizontal stripes, and our featurs are composed of color features including RGB, HSV, LAB, XYZ, YCbCr and NTSC and texture features including Gabor, Schmid and LBP.
A 16D histogram is extracted for each channel and then normalized by $L_1$-norm.
All histograms are concatenated together to form a single vector.
In this work, we denote the above type of hand-crafted features as ELF16.

\subsection{Proposed New Features}
We aim to jointly map C\nolinebreak[4]NN features and hand-crafted featues to a unitary feature space.
A feature fusion deep neural network is proposed in order to use hand-crafted features to regularize CNN features so as to make CNN extract complementary features. In our framework, by using back propagation, the parameters of the whole CNN network could be affected by hand-craft features.
In general, as a results of the fusion, the regularized CNN features output by our proposal network should be more discriminative than both CNN features and the employed hand-crafted features.



\vspace{0.2cm}

\noindent {\bf Fusion Layer and Buffer Layer.} Our Fusion Layer uses full connection to provide self-adaptation on person re-identification problems.\
Both ELF16 Features and CNN Features are followed by a 4096D-output full connection layer (Buffer Layer), which provides buffer for the fusion action.
Buffer Layer is essential in our architecture, since it bridges the gap between two features with huge difference, and guarantees the convergence of FFN.

If the input of Fusion Layer is
\begin{equation}
\bm{x}~=~[\bm{ELF16}, \bm{CNN\_Features}],
\end{equation}
then the output of this layer is computed by:
\begin{equation}
\bm{Z}_{Fusion}(\bm{x})=h(\bm{W}_{Fusion}^T\bm{x}+\bm{b}_{Fusion}),
\end{equation}
where $h(\cdot)$ denotes the activation function. The ReLU and dropout layers are adopted, with a dropout ratio $0.5$.
According to back propagation algorithm, parameters of $l^{th}$ layer after a new iteration are written as:
\begin{equation}
\bm{W}^{(l)}_{new}=\bm{W}^{(l)}-\alpha[(\frac{1}{m}\Delta \bm{W}^{(l)})+\lambda \bm{W}^{(l)}],
\end{equation}
\begin{equation}
\bm{b}^{(l)}_{new}=\bm{b}^{(l)}-\alpha[\frac{1}{m}\Delta \bm{b}^{(l)}],
\end{equation}
where parameters $\alpha$, $m$ and $\lambda$ are set under the guidance of ~\cite{bottou2012stochastic}.

Existing deep re-identification networks for person re-identification adopt Deviance Loss~\cite{yi2014deep} or Maximum Mean Discrepancy~\cite{ahmed2015improved} as loss function.
But we aim at extracting deep features on every image effective\nolinebreak[4]ly rather than performing pairwise comparison through a deep neural network.
Therefore, softmax loss function is applied in our model, and intuitively speaking a more discriminative feature representation should result in lower softmax loss as well.
 For a single input vector $\bm{x}$ and a single output node $j$ in the last layer, the loss could be calculated by:
\begin{equation}
p(y=j|\bm{x};\bm{\theta}) = \frac{e^{\bm{\theta}_j^T\bm{x}}}{\sum_{k=1}^n e^{\bm{\theta}^T_k\bm{x}}}.
\end{equation}
The last layer of our network is designed to minimize the cross-entropy loss:
\begin{equation}
J=-\sum_{k=1}^n \bm{p}_k\log \bm{p}_k,
\end{equation}
in which the number of output node $n$ varies on different training sets as described in Section~\ref{section:training strategy}.

\subsection{How do Hand-crafted Features Influence the Extraction of CNN Features?}
If the parameters of the network are influenced by the ELF16 features $\bm{\tilde{x}}$, \ie, the gradient of the network parameters are adjusted according to $\bm{\tilde{x}}$, then ELF16 features in the lower part of FFN could make CNN features more complementary with it, since the final objective of FFN is to make our features more discriminative in different images.

Denote CNN features (in FC7 layer) as $\bm{x}$ and ELF16 features as $\bm{\tilde{x}}$, Denote the weight connecting the $j^{th}$ node in  $n^{th}$ layer and the $i^{th}$ node in $(n+1)^{th}$ layer as $\bm{W}_{ij}^n$. \  Let $\bm{Z}^n_j = \sum_j \bm{W}^{n-1}_{ji} \bm{a}_i^{n-1}$ where $\bm{a}_i^{n-1} = h(\bm{Z}_i^{n-1})$.
Denote
\begin{equation}
\bm{\delta}_i^n = \frac{\partial J}{\partial \bm{Z}_i^n}.
\end{equation}
Note that $\bm{Z}^8_j = \sum_j \bm{W}^{n-1} \bm{x}_i^{n-1}$. We show that by using back propagation, $\frac{\partial \bm{J}}{\partial \bm{W}_{ij}^7}$ is influenced by $\bm{\tilde{x}}$. In this way, CNN Features learn its parameters which will form features complementary to the ELF16 features $\bm{\tilde{x}}$.
Note that
\begin{equation}\label{why1}
  \frac{\partial J}{\partial \bm{W}_{ij}^7} = \bm{x}_j \bm{\delta}_i^8,
\end{equation}
where
\begin{equation}\label{why2}
  \bm{\delta}_i^8 = (\sum_{j} \bm{W}_{ji}^8 \bm{\delta}_j^9) h'(\bm{Z}_i^8),
\end{equation}
\begin{equation}\label{why3}
  \bm{\delta}_j^9 = (\sum_{k} \bm{W}_{kj}^9 \bm{\delta}_k^{10}) h'(\bm{Z}_j^9).
\end{equation}
$\bm{\delta}_j^9$ is influenced by $\bm{\tilde{x}}$ in two ways.
Firstly,
\begin{equation}\label{why4}
  \bm{Z}_k^9 = \sum_j \bm{W}_{kj}^8 \bm{a}^8_j + \sum_j \bm{\tilde{W}}_{kj}^8 \bm{\tilde{a}}^8_j,
\end{equation}
where
\begin{equation}\label{why5}
  \bm{\tilde{a}}^8_j = h(\sum_i \bm{\tilde{W}}_{ji}^7 \bm{\tilde{x}}_i).
\end{equation}
In other words, the information in ELF16 features $\bm{\tilde{x}}$ could propagate through $h'(\bm{Z}_j^9)$, and thus the convolution filters of Deep Feature Extraction part would adapt itself according to $\bm{\tilde{x}}$.
Secondly, the output of softmax loss layer is influenced by $\bm{\tilde{x}}$ during the forward propagation process, and thus $ \bm{\delta}_k^{10}$ is also influenced by $\bm{\tilde{x}}$.

\section{Settings for Feature Fusion Network}
\label{section:training strategy}
\subsection{Training Dataset}
Market-1501 is a multi-shot person re-identification dataset recently reported by \cite{zheng2015scalable}.
It consists of 38195 images from 1501 identities, which is the largest public person re-identification dataset available.
We trained our Feature Fusion Network on Market-1501, and used it to extract features in Section~\ref{section:experiments}.
\subsection{Training Strategies}
\begin{figure*}[htbp]
\centering {
\subfigure[ VIPeR on $L_1-norm$ ]
{
    \label{fig: viper to ilids}
    \includegraphics[width=0.315\linewidth,height=0.2\linewidth]{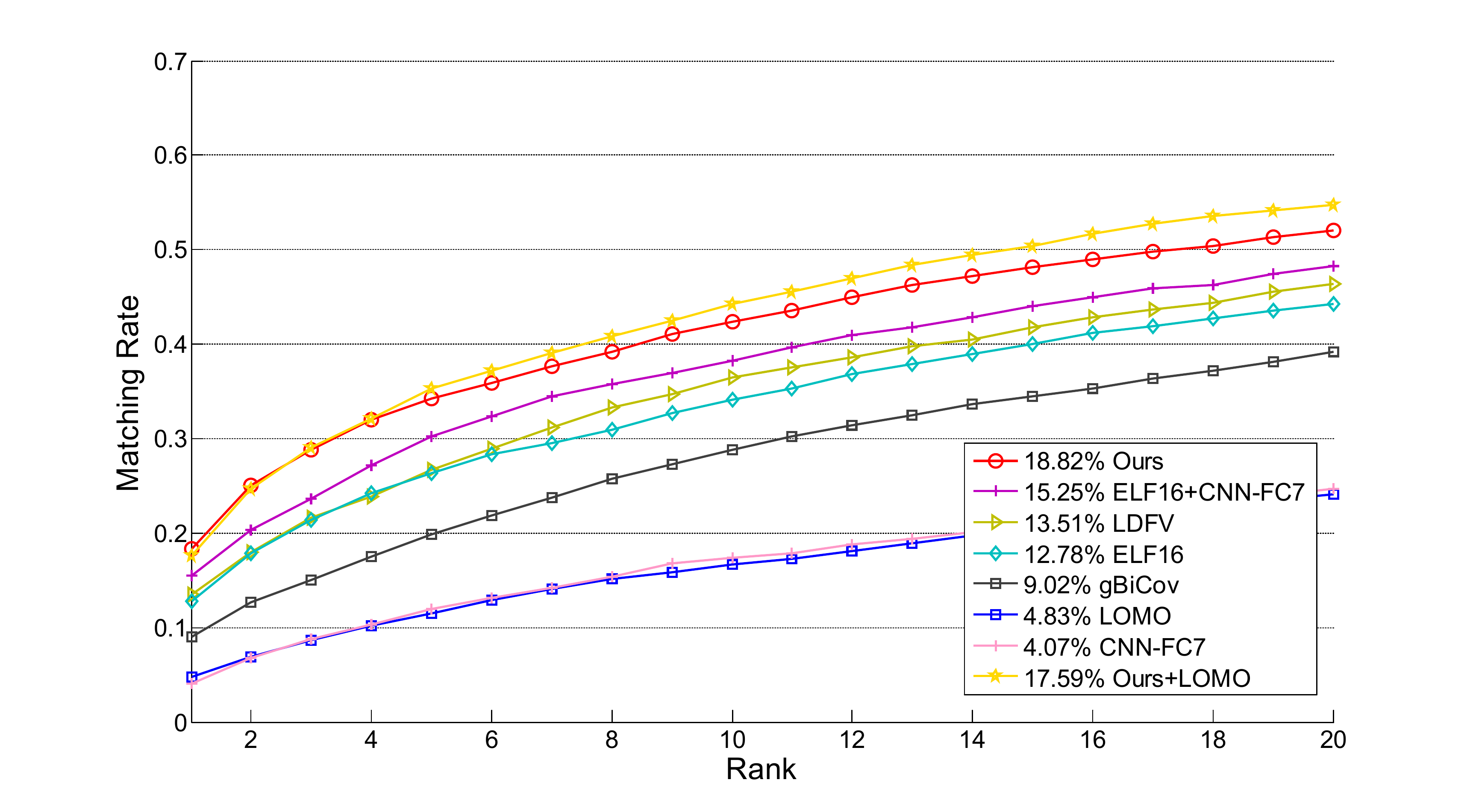}
}
\subfigure[ CUHK01 on $L_1-norm$ ]
{
    \label{fig: 3dpes to ilids}
    \includegraphics[width=0.315\linewidth,height=0.2\linewidth]{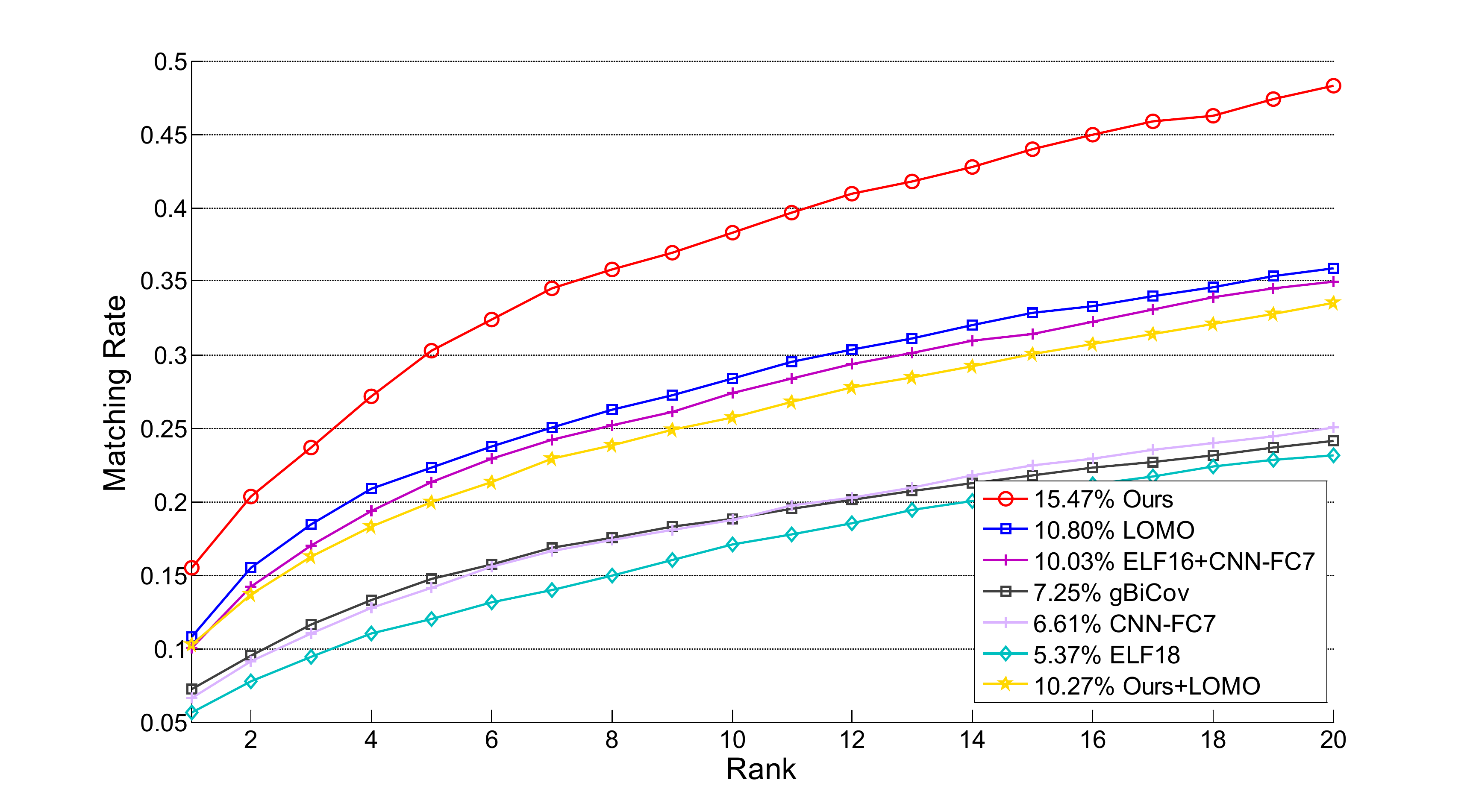}
}
\subfigure[ PRID450s on $L_1-norm$ ]
{
    \label{fig: cavidara to ilids}
    \includegraphics[width=0.315\linewidth,height=0.2\linewidth]{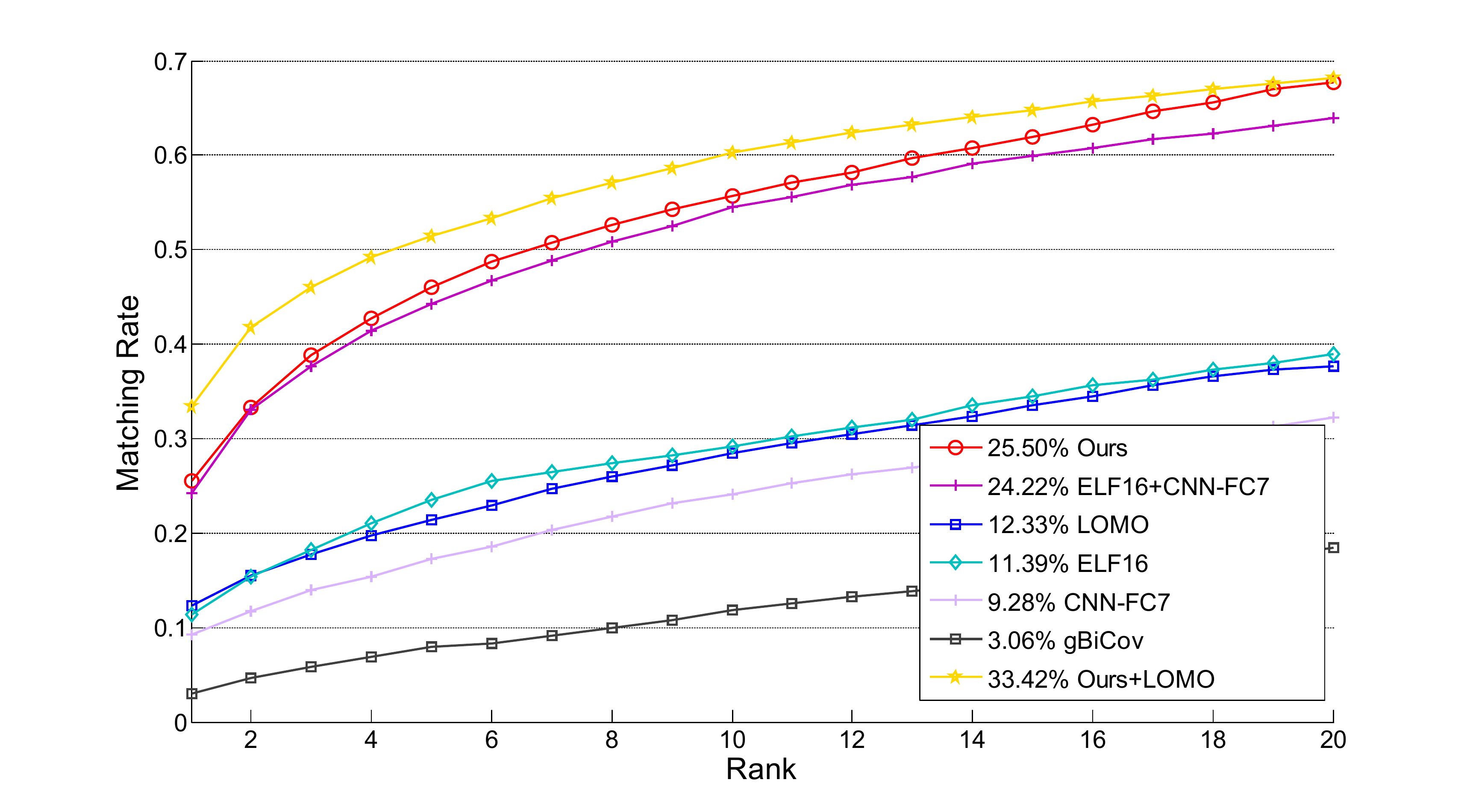}
}
}
\centering{
\subfigure[ VIPeR on LFDA ]
{
    \label{fig: viper to cavidara}
    \includegraphics[width=0.315\linewidth,height=0.2\linewidth]{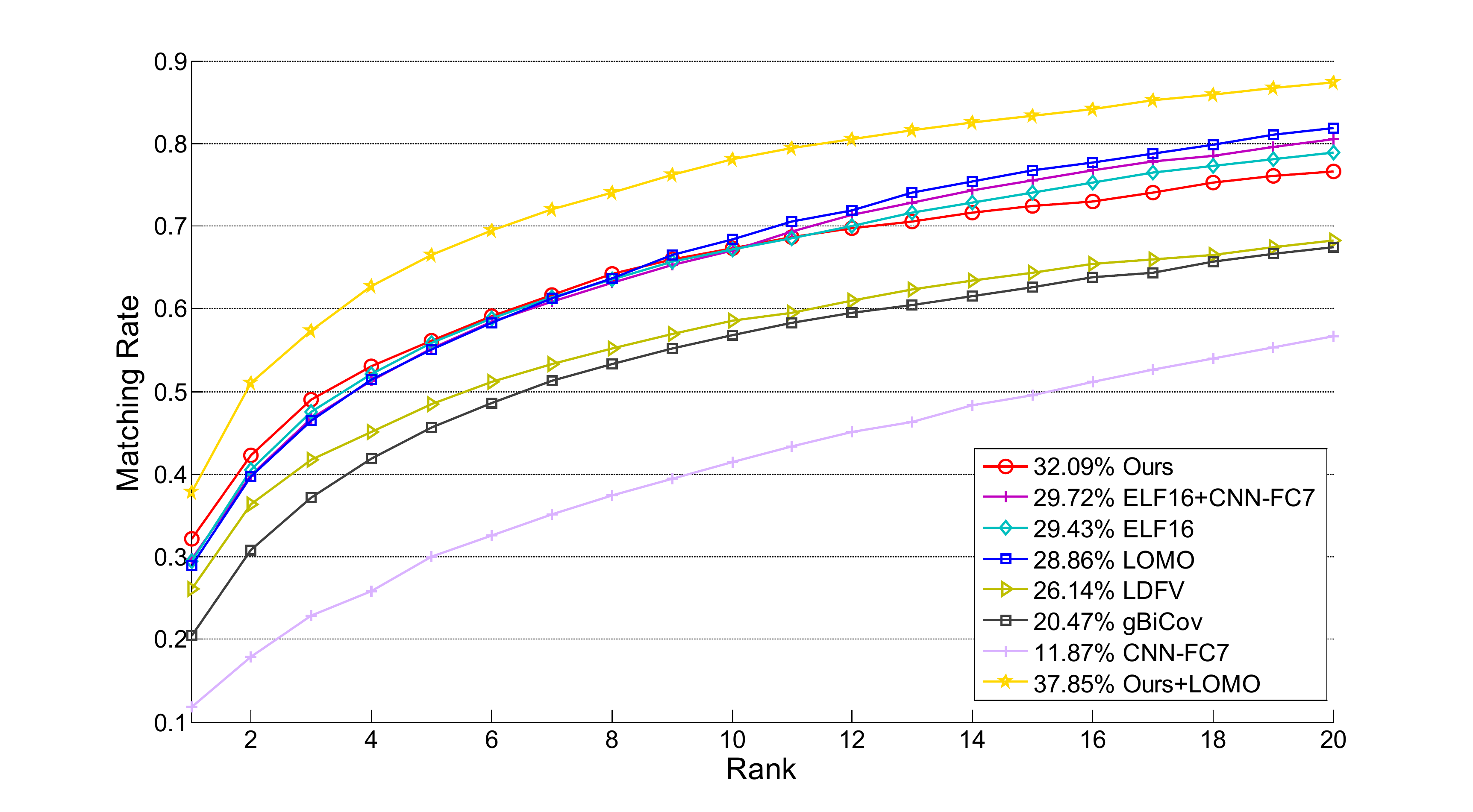}
}
\subfigure[ CUHK01 on LFDA ]
{
    \label{fig: 3dpes to cavidara}
   \includegraphics[width=0.315\linewidth,height=0.2\linewidth]{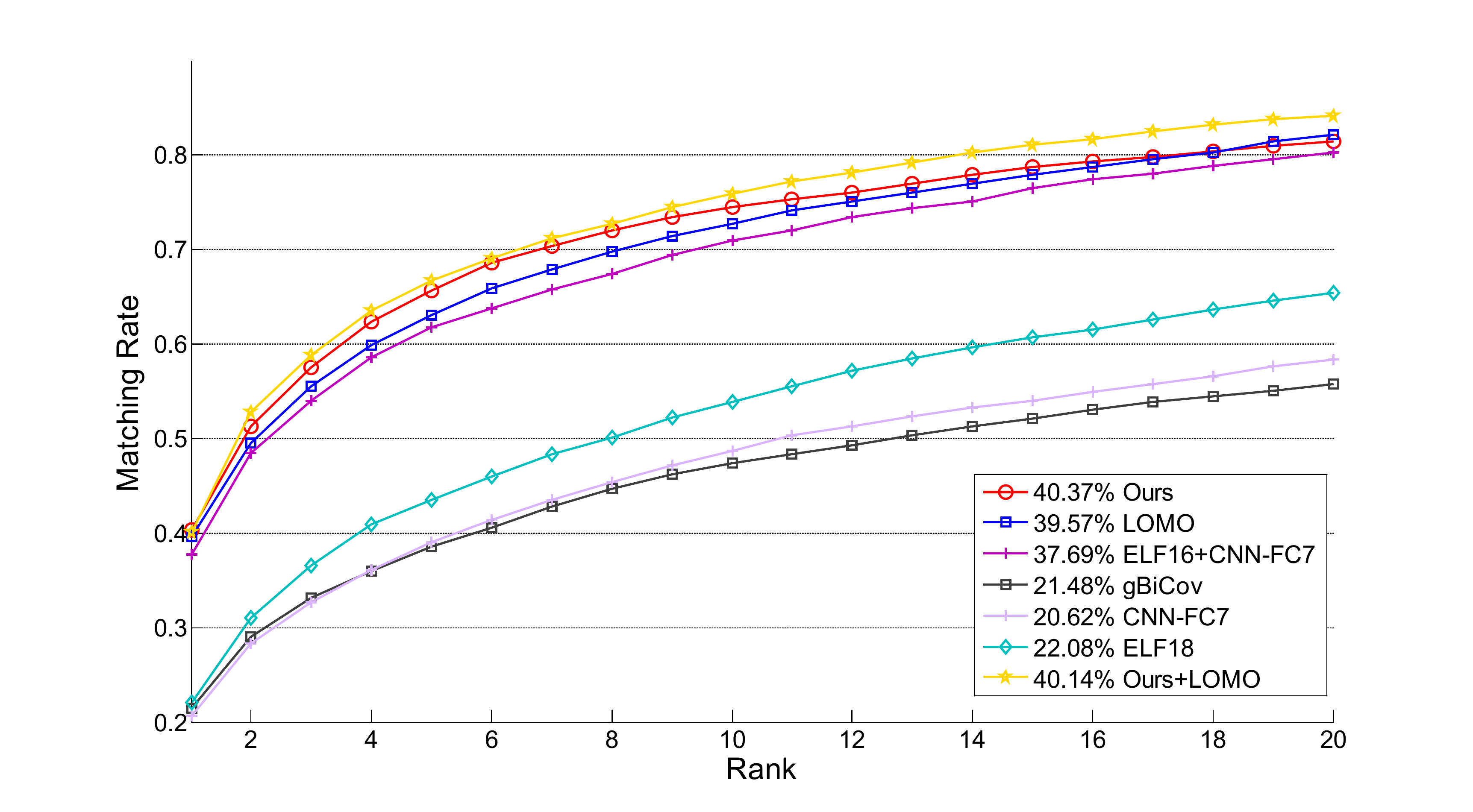}
}
\subfigure[ PRID450s on LFDA ]
{
    \label{fig: ilids to cavidara}
    \includegraphics[width=0.315\linewidth,height=0.2\linewidth]{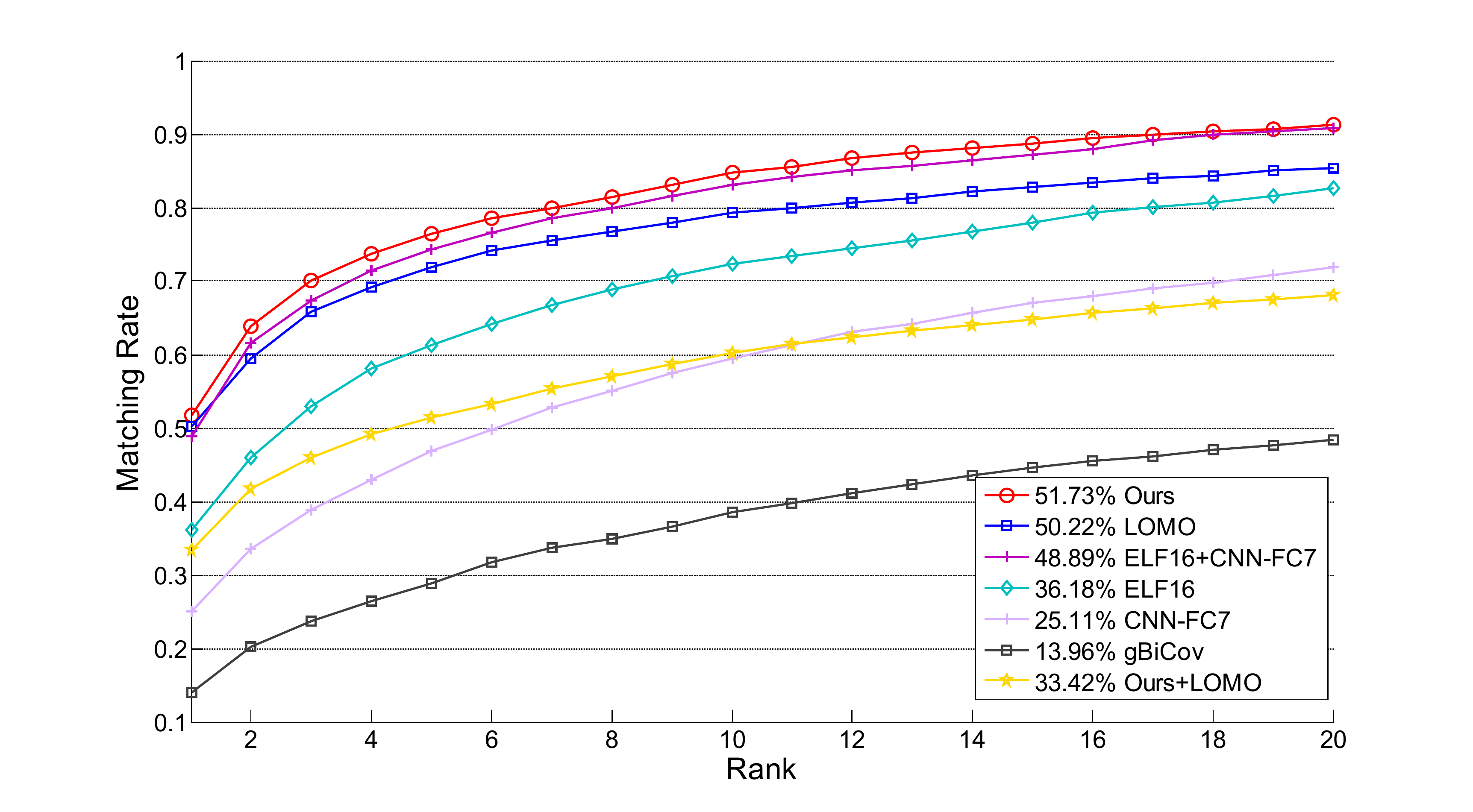}
}
}
\centering{
\subfigure[VIPeR on Mirror KMFA ]
{
    \label{fig: viper to 3dpes}
    \includegraphics[width=0.315\linewidth,height=0.2\linewidth]{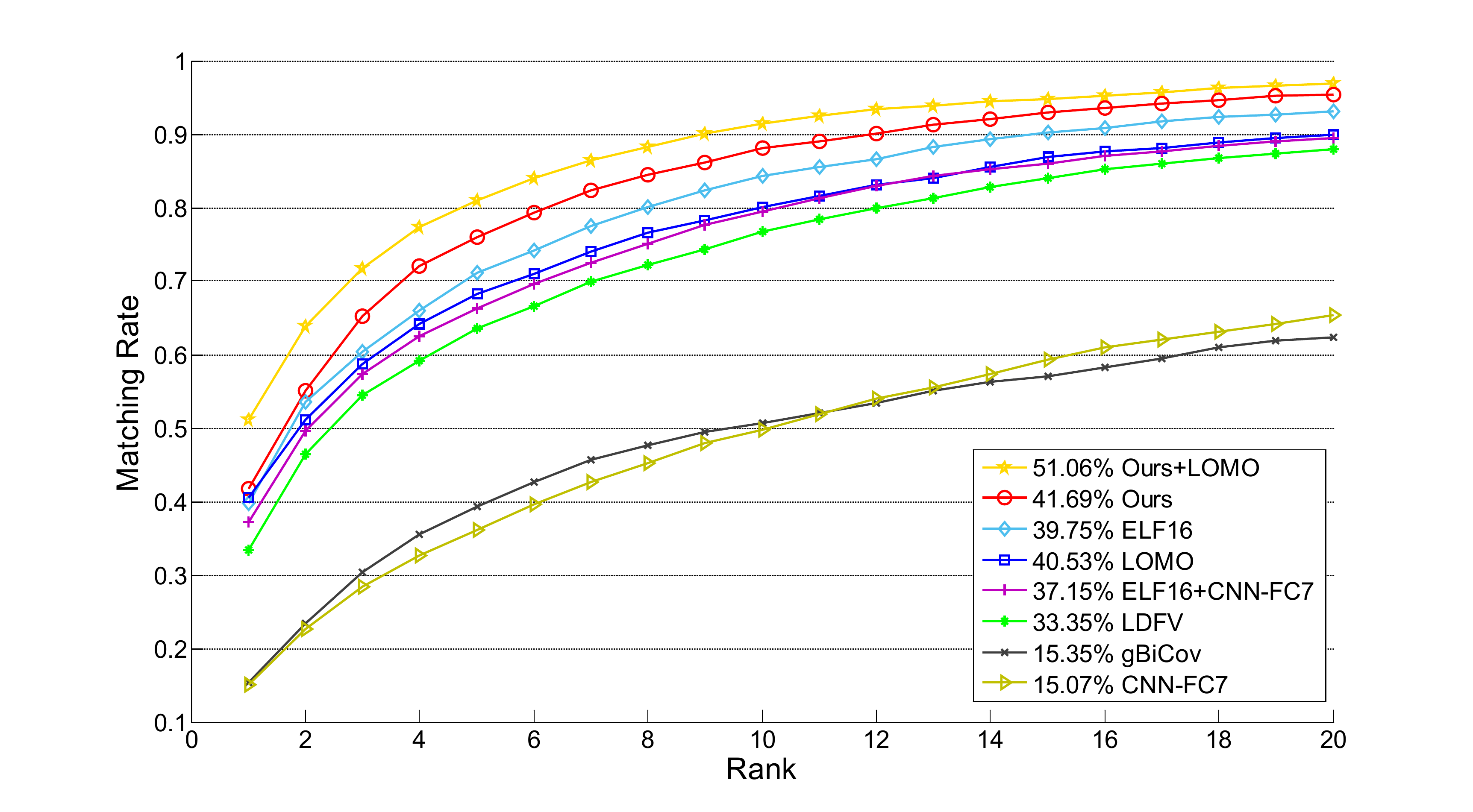}
}
\subfigure[CUHK01 on Mirror KMFA ]
{
    \label{fig: ilids to 3dpes}
    \includegraphics[width=0.315\linewidth,height=0.2\linewidth]{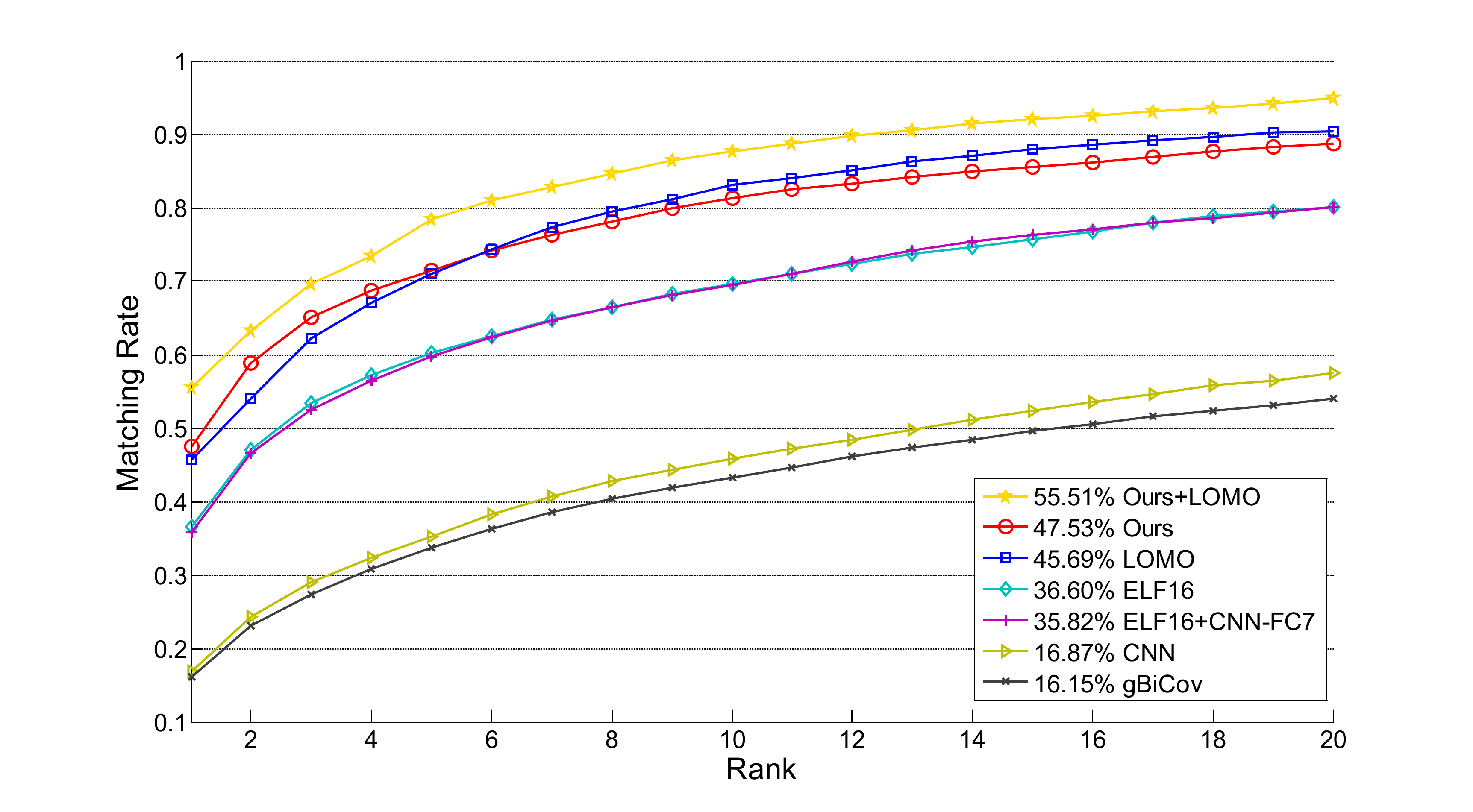}
}
\subfigure[PRID450s on Mirror KMFA ]
{
    \label{fig: cavidara to 3dpes}
   \includegraphics[width=0.315\linewidth,height=0.2\linewidth]{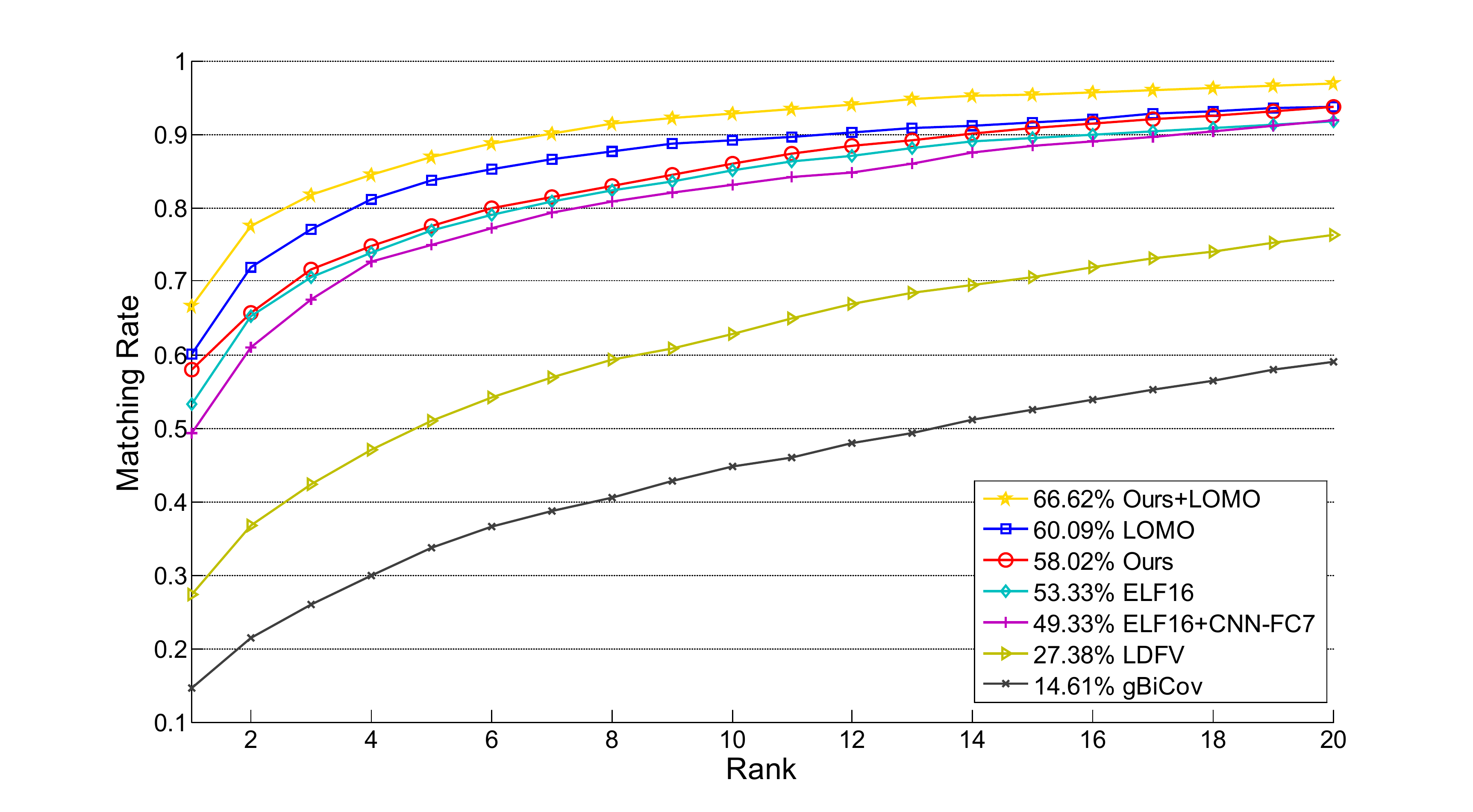}
}
}
\caption{CMC Curves of three datasets. $L_1-norm$, LFDA and Mirror KMFA were used to evaluate the features.
The yellow CMC Curves in the last row indicates the final model we used in Section~\ref{section:state of the art}.}\label{fig: methods comparison}
\label{fig: comparison on different classifiers}
\end{figure*}
Our training strategy applied mini-batch stochastic gradient descent (SGD) for faster back propagation and smoother convergence~\cite{bottou2012stochastic}.
In each iteration of training phase, 25 images form a mini-batch and were forwarded to softmax loss Layer.
The initial learning rate $\gamma = 1 e -5$, which is significantly smaller than most of other CNN m\nolinebreak[4]odel\nolinebreak[4]s.
Ev\nolinebreak[4]ery 20000 iterations the learning rate decreased by $\gamma_{new}=0.1*\gamma$.
We finetuned our network based on ImageNet~\cite{krizhevsky2012imagenet} model provided by~\cite{jia2014caffe}.
Our FFN model took 50000 iterations to converge (about 4 hours on a Tesla K20m GPU).
In order to improve the adaptation of our model, we further use difficult samples to finetune the network.

Hard negative mining~\cite{ahmed2015improved} gives us a logical way to emphasize difficult samples in CNN.
This training strategy is originally designed to balance the positive and negative samples in pairwise comparison for person re-identification.
We applied this strategy to our Feature Fusion Network as well.
About 12000 images of 630 IDs were wrongly-labeled by the previous network, and were manually picked out for further finetuning.
We replaced the last softmax loss layer with less output nodes and continued to finetune our model on these difficult samples, with lower learning rate ($1 e -6$) and fewer iterations (about 10000).
The whole training process took about 5-6 hours to converge to a tolerable training loss (about 0.05 typically).

\section{Experiments}
\label{section:experiments}
This section evaluated our new features in different perspectives.
We presented extensive experimental results on three benchmark datasets in order to clearly demonstrate the effectiveness of our features.

\subsection{Datasets and Experiment Protocols}
Our test was based on three publicly available datasets: VIPeR~\cite{gray2007evaluating}, CUHK01~\cite{li2012human} and PRID450s~\cite{roth2014mahalanobis}.
Each of our datasets was presented in two disjoint camera views, with significant misalignment, light change and body part distortion.
Table~\ref{Datasets} briefly introduces these three datasets.
Also, some sample images of these datasets are shown in Fig.~\ref{fig: datasets_demo}.

In each individual experiment, we randomly selected half of the identities as training set, and the other half as testing set.
Training set was used to train projection matrix $\bm{W}$ (in metric learning methods).
Testing set used $\bm{x}'=\bm{W}^T\bm{x}$ to get the final projection of $\bm{x}$ and measures the distance between a pair of input images.
For the reliability and stableness of our results, each experiment was repeated 10 times and the average Rank-i accuracy rate was computed.
Cumulative Matching Curve (CMC) was also provided in Fig.~\ref{fig: methods comparison}, providing a more intuitional comparison between different algorithms.

We applied single-shot protocol in our experiment, that is dur\nolinebreak[4]ing testing phase, one image was chosen from View2 as probe and all the images in View1 were regarded as the gallery.
For CUHK01 specifically, which has 2 images of the same person in one camera view, we randomly chose one image of each identity as the gallery.

Mirror Kernel Marginal Analysis (KMFA), proposed by~\cite{chen2015mirror}, provides a high-performance metric learning algorithm on person re-identification.
This method was adopted in Section~\ref{section:KMFA}, with chi-square kernel embedded and parameters set to the optimal according to ~\cite{chen2015mirror}.

\subsection{Features}
Six feature extraction approaches were evaluted in our experiments for comparison, including LDFV~\cite{ma2012local}, gBiCov~\cite{ma2014covariance},
ImageNet~\cite{krizhevsky2012imagenet} CNN features, LOMO features~\cite{liao2015person}, ELF16 features and our proposed features\footnote{Our proposed features are available at http://isee.sysu.edu.cn/resource}.
For ImageNet CNN features alone, FC7 layer data, which produced the highest accuracy rate in our tests, was chosen in our experiments.
Lo\nolinebreak[4]cal Maximal Occurrence Representation (LOMO) is another high-performance feature representation specifically designed for re-identification problem.
LDFV features were evaluated only on VIPeR dataset due to its copyrights of code.
All images were resized to 224$\times$224 for our feature extraction.

In order to demonstrate the effectiveness of our new feature, two compound features (ELF16+CNN-FC7 and Ours+LOMO) were also added for the comparison.
ELF16+CNN-FC7 denotes the concatenation of normalized CNN-FC7 feature to ELF16 feature.
Ours+LOMO denotes the concatenation of our new features and normalized LO\nolinebreak[4]MO features.

All of these features were extracted and evaluated in its default dimension (see Table~\ref{cost of time}).
\begin{table}
\footnotesize
\begin{center}
\setlength{\tabcolsep}{6pt}
\begin{tabular}{|c|c|c|c|}
\hline
&VIPeR & CUHK01 & PRID450s \\
\hline
No. of images & 1264 & 3884&900 \\
\hline
No. of identities & 632&971&450 \\
\hline
No. of images in training set& 316&485 & 225 \\
\hline
No. of camera views& 2&2 & 2 \\
\hline
No. of images per view per ID&1&2&1\\
\hline
\end{tabular}
\end{center}
\caption{Re-identification datasets used in our experiments.}
\label{Datasets}
\end{table}

\subsection{Evaluations on Features}
\subsubsection{Unsupervised Method}
~~~~Fig.~\ref{fig: methods comparison}~(a)-(c) shows the performance of our features compared to other features on $L_1-norm$, evaluating an algorithm's capability in an original and unsupervised perspective.
Our features significantly outperformed other stand-alone features (see Fig.\ref{fig: methods comparison}~(a)-(c)), suggesting that raw information provided by our feature is more accurate for representing re-identification images in most cases.

ELF16+CNN-FC7 features performed the second-best and outperformed both ELF16 and CNN-FC7, which provides supports on our assumption that traditional feature and CNN features are complementary.
 Also, our new features significantly outperformed ELF16+CNN-FC7, which may be bause of the following two reasons:
\begin{itemize}
\item CNN features in our network were trained to be complementary to the traditional features, while in ELF16+CNN-FC7, the CNN features are simply cascad\nolinebreak[4]ed with ELF16 features, which may not be optimal.
\item The use of Buffer Layer and Fusion layer could automatically tune the weights for each feature, and makes the fused feature perform much better.
\end{itemize}

LOMO features were specifically designed to describe person re-identification images.
However, it ranked the seventh on VIPeR and the third on CUHK01, which is not stable enough for $L_1-norm$.

\subsubsection{Metric Learning Methods}
\label{section:KMFA}
~~~~To demonstrate the maximal effectiveness of our im\nolinebreak[4]age description, we put it into two metric learning methods: LF\nolinebreak[4]DA~\cite{pedagadi2013local} and Mirror KMFA~\cite{chen2015mirror}, along with other widely-used features.
 We used each of the features to learn distance metric between each probe image and gallery set.
In this experiments, we evaluated their capability on supervised metric learning methods.

Fig.~\ref{fig: comparison on different classifiers} (d)-(i) shows the CMC curves on three dataset\nolinebreak[4]s, with Rank-1 identification rate labeled on each feature type.
Note that LDFV performed badly using chi-square kernel, so we adopted Mirror MFA without kernel trick in the comparison.

The results clearly show the outstanding performance of our proposed features, as it exceeded all the stand-alone features in VIPeR and CUHK01.
Compared to ELF16 and CNN-FC7 features alone, our new features yielded much better results.
Also, the simple concatenation of these two features (ELF+CNN-FC7) could not represent the image as good as ours, and it indicates the necessity of Fusion Layer in the proposed FFN.
\begin{table}[htbp]
\footnotesize
\begin{center}
\setlength{\tabcolsep}{8pt}
\begin{tabular}{|c|c|c|c|c|c|}
\hline
 Rank &  1&5&10&20\\
\hline
Our Model&\textbf{51.06}&\textbf{81.01}&\textbf{91.39}&\textbf{96.90}\\
\hline
Deep Feature Learning\cite{ding2015deep}&40.50&60.80&70.40&84.40\\
\hline
LOMO+XQDA~\cite{liao2015person}&40.00&67.40&80.51&91.08\\
\hline
Mirror KMFA($R_{\chi^2}$)~\cite{chen2015mirror}&42.97&75.82&87.28&94.84\\
\hline
mFilter+LADF~\cite{zhao2014learning}&43.39&73.04&84.87&93.70\\
\hline
mFilter~\cite{zhao2014learning}&29.11&52.10&67.20&80.14\\
\hline
SalMatch~\cite{zhao2013person}&30.16&52.31&65.54&79.15\\
\hline
LFDA~\cite{pedagadi2013local}&24.18&52.85&67.12&78.96\\
\hline
LADF~\cite{li2013learning}&29.34&61.04&75.98&88.10\\
\hline
RDC~\cite{zheng2013reidentification}&15.66&38.42&53.86&70.09\\
\hline
KISSME~\cite{koestinger2012large}&24.75&53.48&67.44&80.92\\
\hline
LMNN-R~\cite{dikmen2011pedestrian}&19.28&48.71&65.49&78.34\\
\hline
PCCA~\cite{mignon2012pcca}&19.28&48.89&64.91&80.28\\
\hline
$L_2-norm$&10.89&22.37&32.34&45.19\\
\hline
$L_1-norm$&12.15&26.01&32.09&34.72\\
\hline
\end{tabular}
\end{center}
\caption{Top Matching Rank on VIPeR (Sorted by the proposed time).}
\label{Top Matching Rank on VIPeR.}
\end{table}
\begin{table}[htbp]
\footnotesize
\begin{center}
 \setlength{\tabcolsep}{8pt}
\begin{tabular}{|c|c|c|c|c|c|}
\hline
 Rank &  1&5&10&20\\
\hline
Our Model&\textbf{55.51}&\textbf{78.40}&\textbf{83.68}&\textbf{92.59}\\
\hline
Mirror KMFA($R_{\chi^2}$)~\cite{chen2015mirror}&40.40&64.63&75.34&84.08\\
\hline
Ahmed's Deep Re-id~\cite{ahmed2015improved}&47.53&72.10&80.53&88.49\\
\hline
mFilter~\cite{zhao2014learning}&34.30&55.12&64.91&74.53\\
\hline
SalMatch~\cite{zhao2013person}&28.45&45.85&55.67&67.95\\
\hline
DeepReID~\cite{li2014deepreid}&27.87&64.01&82.50&87.36\\
\hline
ITML~\cite{davis2007information}&15.98&35.22&45.60&59.81\\
\hline
eSDC~\cite{zhao2013unsupervised}&19.67&32.72&40.29&50.58\\
\hline
LFDA~\cite{pedagadi2013local}&22.08&41.56&53.85&64.51\\
\hline
KISSME~\cite{koestinger2012large}&14.02&32.20&44.44&56.61\\
\hline
LMNN-R~\cite{dikmen2011pedestrian}&13.45&31.33&42.25&54.11\\
\hline
$L_2-norm$&5.63&16.00&22.89&30.63\\
\hline
$L_1-norm$&10.80&15.51&37.57&35.57\\
\hline
\end{tabular}
\end{center}
\caption{Top Matching Rank on CUHK01(Sorted by proposed time).}
\label{Top Matching Rank on CUHK01.}
\end{table}
\begin{table}[htbp]
\footnotesize
\begin{center}
 \setlength{\tabcolsep}{8pt}
\begin{tabular}{|c|c|c|c|c|c|}
\hline
 Rank &  1&5&10&20\\
\hline
Our Model&\textbf{66.62}&\textbf{86.84}&\textbf{92.84}&\textbf{96.89}\\
\hline
Mirror KMFA($R_{\chi^2}$)~\cite{chen2015mirror}&55.42&79.29&87.82&93.87\\
\hline
Ahmed's Deep Re-id~\cite{ahmed2015improved}&34.81&63.72&76.24&81.90\\
\hline
ITML~\cite{davis2007information}&24.27&47.82&58.67&70.89\\
\hline
LFDA~\cite{pedagadi2013local}&36.18&61.33&72.40&82.67\\
\hline
KISSME~\cite{koestinger2012large}&36.31&65.11&75.42&83.69\\
\hline
LMNN-R~\cite{dikmen2011pedestrian}&28.98&55.29&67.64&78.36\\
\hline
$L_2-norm$&11.33&24.50&33.22&43.89\\
\hline
$L_1-norm$&25.50&25.33&51.73&53.07\\
\hline
\end{tabular}
\end{center}
\caption{Top Matching Rank on PRID450s (Sorted by proposed time)}
\label{Top Matching Rank on PRID450s.}
\end{table}

Our proosed features are always better than LOMO features.
Since LOMO emphasizes on HSV and SILT\nolinebreak[4]P histograms, it performed better on PRID450s, which was undergoing specific lighting conditions. But on other datasets, our new features are still better than LOMO features.

The concatenation of these two features (Ours+LOMO) has a strong discriminative ability and outperformed all other features on Mirror KMFA.
This indicates that our deep learning methods is complementary to LOMO features.
Thus, we regard this 31056D mix features as the final image representation in Mirror KMFA person re-identification model.

\subsection{Comparison with State-of-the-Art}
\label{section:state of the art}
This experiment compared overall performance between state-of-the-art person re-identification model and ours.
Our model is based on Mirror KMFA, using the concatenation of our new features and normalized LOMO features (Ours+LOMO).

Table~\ref{Top Matching Rank on VIPeR.}-\ref{Top Matching Rank on PRID450s.} summarize some of the highest performance models on VIPeR, CUHK01 and PRID450s, including
LOMO+XQDA~\cite{liao2015person}, Mirror KMFA~\cite{chen2015mirror}, Ahmed's Improved Deep ReID~\cite{ahmed2015improved} and Mid-level Filter~\cite{zhao2014learning}.
Our model can beat them by about 10\% in Rank-1 matching rate.

Three Deep Learning methods (DeepReID~\cite{li2014deepreid}, Ahmed's Deep Re-id~\cite{ahmed2015improved}, Ding's Deep Feature Learning~\cite{ding2015deep})are specifically listed in Table~\ref{Top Matching Rank on CUHK01.} and~\ref{Top Matching Rank on PRID450s.}.
All of them modified CNN for pairwise comparison, and employed unique layers to match two views of the input images.

In comparison, our model regards CNN as a feature extractor, while adopting metric learning to calculate relative distance of different images.
This not only contributes to the improvement in accuracy, but also enables us to use larger datasets in CNN training process.
Our model also clearly exceeded their performance on CUHK01 (7.98\%) and PRID450s (11.2\%).
\subsection{Running Time}
 We evaluate the running time of these feature-extraction algorithms, as shown in Fig.~\ref{cost of time}.
The reporeted time is the average feature extraction time for a single $48\times 128$ image on VIPeR dataset (in its default dimension).
 Note that we have included the time of extracting ELF16 features in the last row.
 bottou2012stochastic
 It can be seen that our Fusion Feature Network is even faster than some of the hand-crafted methods (such as gBiCov), which breaks the stereotype of huge and clumsy Convolutional Neural Network.
 Also, most of the time was spent on the extraction of ELF16 features.
 Compared to LOMO features, our features have much lower dimension, and will perform faster in the metric learning step followed.
 With a balance between the speed and dimensional complexity, our Feature Fusion Network can be easily applied to actual use.
 Besides, compared to other CNN-based models, our FFN does not need to finetune on target datasets, which makes it faster to apply.
\renewcommand{\multirowsetup}{\centering}
\begin{table}
\footnotesize
\begin{center}
 \setlength{\tabcolsep}{8pt}
\begin{tabular}{|c|c|c|}
\hline
Feature & Extraction Time & Default Dimension \\
\hline
gBiCov & 13.6152s & 5940 \\
\hline
LOMO & 0.2610s & 26960 \\
\hline
ELF16 & 0.5720s & 8064 \\
\hline
CNN-FC7 & 0.1773s & 4096 \\
\hline
Ours (with ELF16) &0.1769s+0.5720s&4096\\
\hline
\end{tabular}
\end{center}
\caption{Average time of extracting features of a single 64x128 image (Evaluated on a 2.00GHz Xeon CPU with 16 cores).}
\label{cost of time}
\end{table}

\section{Conclusion}
In this paper, we have presented a novel and effective way of feature extraction for person re-identification called Feature Fusion Network (FFN).
This model jointly utilizes both CNN feature and hand-crafted features, including RG\nolinebreak[4]B, HSV, YCbCr, Lab, YIQ color feature and Gabor texture feature.
It could adjust the weights of these information automatically with the back propagation process of Neural Network.
Also, we have proved that FFN regularizes the CNN process so as to make CNN focus on extracting complementary features.
Experiments on three challenging person re-identification datasets (VIPeR, CUHK01, PRID450s) show the effectiveness of our learned deep features.
By using Mirror Kernel Marginal Fisher Analysis (KMFA), our proposed features significantly outperform the state-of-the-art person re-identification models on these three datasets by 8.09\%, 7.98\%, and 11.2\% (in Rank-1 accuracy rate), respectively.

\section*{Acknowledgements}

This research was partly supported by Guangdong Provincial Government
of China through the Computational Science Innovative Research Team
Program, and partially by Natural Science Foundation of China (Nos.
61472456, 61522115, 61573387), Guangzhou Pearl River
Science and Technology Rising Star Project under Grant 2013J2200068, the
Guangdong Natural Science Funds for Distinguished Young Scholar under
Grant S2013050014265, and the GuangDong Program (No. 2015B010105005).
{\small
\bibliographystyle{ieee}
\bibliography{egbib}

\begin{thebibliography}{10}\itemsep=-1pt

\bibitem{ahmed2015improved}
E.~Ahmed, M.~Jones, and T.~K. Marks.
\newblock An improved deep learning architecture for person re-identification.
\newblock In {\em IEEE CVPR}, 2015.

\bibitem{bottou2012stochastic}
L.~Bottou.
\newblock Stochastic gradient descent tricks.
\newblock In {\em Neural Networks: Tricks of the Trade}. 2012.

\bibitem{chen2015mirror}
Y.-C. Chen, W.-S. Zheng, and J.~Lai.
\newblock Mirror representation for modeling view-specific transform in person
  re-identification.
\newblock In {\em IJCAI}, 2015.

\bibitem{davis2007information}
J.~V. Davis, B.~Kulis, P.~Jain, S.~Sra, and I.~S. Dhillon.
\newblock Information-theoretic metric learning.
\newblock In {\em ICML}, 2007.

\bibitem{dikmen2011pedestrian}
M.~Dikmen, E.~Akbas, T.~S. Huang, and N.~Ahuja.
\newblock Pedestrian recognition with a learned metric.
\newblock In {\em ACCV}. 2011.

\bibitem{ding2015deep}
S.~Ding, L.~Lin, G.~Wang, and H.~Chao.
\newblock Deep feature learning with relative distance comparison for person
  re-identification.
\newblock {\em Pattern Recognition}, 2015.

\bibitem{farenzena2010person}
M.~Farenzena, L.~Bazzani, A.~Perina, V.~Murino, and M.~Cristani.
\newblock Person re-identification by symmetry-driven accumulation of local
  features.
\newblock In {\em IEEE CVPR}, 2010.

\bibitem{gray2007evaluating}
D.~Gray, S.~Brennan, and H.~Tao.
\newblock Evaluating appearance models for recognition, reacquisition, and
  tracking.
\newblock In {\em IEEE PETS Workshop}, 2007.

\bibitem{gray2008viewpoint}
D.~Gray and H.~Tao.
\newblock Viewpoint invariant pedestrian recognition with an ensemble of
  localized features.
\newblock In {\em ECCV}. 2008.

\bibitem{hu2015deep}
J.~Hu, J.~Lu, and Y.-P. Tan.
\newblock Deep transfer metric learning.
\newblock In {\em IEEE CVPR}, 2015.

\bibitem{jia2014caffe}
Y.~Jia, E.~Shelhamer, J.~Donahue, S.~Karayev, J.~Long, R.~Girshick,
  S.~Guadarrama, and T.~Darrell.
\newblock Caffe: Convolutional architecture for fast feature embedding.
\newblock {\em arXiv:1408.5093}, 2014.

\bibitem{koestinger2012large}
M.~Koestinger, M.~Hirzer, P.~Wohlhart, P.~M. Roth, and H.~Bischof.
\newblock Large scale metric learning from equivalence constraints.
\newblock In {\em IEEE CVPR}, 2012.

\bibitem{krizhevsky2012imagenet}
A.~Krizhevsky, I.~Sutskever, and G.~E. Hinton.
\newblock Imagenet classification with deep convolutional neural networks.
\newblock In {\em NIPS}, 2012.

\bibitem{kviatkovsky2013color}
I.~Kviatkovsky, A.~Adam, and E.~Rivlin.
\newblock Color invariants for person reidentification.
\newblock {\em IEEE TPAMI}, 35(7):1622--1634, 2013.

\bibitem{li2012human}
W.~Li, R.~Zhao, and X.~Wang.
\newblock Human reidentification with transferred metric learning.
\newblock In {\em ACCV}, 2012.

\bibitem{li2014deepreid}
W.~Li, R.~Zhao, T.~Xiao, and X.~Wang.
\newblock Deepreid: Deep filter pairing neural network for person
  re-identification.
\newblock In {\em IEEE CVPR}, 2014.

\bibitem{li2013learning}
Z.~Li, S.~Chang, F.~Liang, T.~S. Huang, L.~Cao, and J.~R. Smith.
\newblock Learning locally-adaptive decision functions for person verification.
\newblock In {\em IEEE CVPR}, 2013.

\bibitem{liao2015person}
S.~Liao, Y.~Hu, X.~Zhu, and S.~Z. Li.
\newblock Person re-identification by local maximal occurrence representation
  and metric learning.
\newblock In {\em IEEE CVPR}, 2015.

\bibitem{ma2012local}
B.~Ma, Y.~Su, and F.~Jurie.
\newblock Local descriptors encoded by fisher vectors for person
  re-identification.
\newblock In {\em ECCV}, 2012.

\bibitem{ma2014covariance}
B.~Ma, Y.~Su, and F.~Jurie.
\newblock Covariance descriptor based on bio-inspired features for person
  re-identification and face verification.
\newblock {\em IVC}, 32(6):379--390, 2014.

\bibitem{mignon2012pcca}
A.~Mignon and F.~Jurie.
\newblock Pcca: A new approach for distance learning from sparse pairwise
  constraints.
\newblock In {\em IEEE CVPR}, 2012.

\bibitem{ojala2002multiresolution}
T.~Ojala, M.~Pietik{\"a}inen, and T.~M{\"a}enp{\"a}{\"a}.
\newblock Multiresolution gray-scale and rotation invariant texture
  classification with local binary patterns.
\newblock {\em IEEE TPAMI}, 24(7):971--987, 2002.

\bibitem{pedagadi2013local}
S.~Pedagadi, J.~Orwell, S.~Velastin, and B.~Boghossian.
\newblock Local fisher discriminant analysis for pedestrian re-identification.
\newblock In {\em IEEE CVPR}, 2013.

\bibitem{roth2014mahalanobis}
P.~M. Roth, M.~Hirzer, M.~K{\"o}stinger, C.~Beleznai, and H.~Bischof.
\newblock Mahalanobis distance learning for person re-identification.
\newblock In {\em Person Re-Identification}, pages 247--267. 2014.

\bibitem{wang2015cross}
X.~Wang, W.-S. Zheng, X.~Li, and J.~Zhang.
\newblock Cross-scenario transfer person re-identification.
\newblock {\em IEEE TCSVT}, PP(99):1--1, 2015.

\bibitem{yi2014deep}
D.~Yi, Z.~Lei, S.~Liao, and S.~Z. Li.
\newblock Deep metric learning for person re-identification.
\newblock In {\em IEEE ICPR}, 2014.

\bibitem{zhang2011gabor}
Y.~Zhang and S.~Li.
\newblock Gabor-lbp based region covariance descriptor for person
  re-identification.
\newblock In {\em IEEE ICIG}, 2011.

\bibitem{zhao2013person}
R.~Zhao, W.~Ouyang, and X.~Wang.
\newblock Person re-identification by salience matching.
\newblock In {\em IEEE ICCV}, 2013.

\bibitem{zhao2013unsupervised}
R.~Zhao, W.~Ouyang, and X.~Wang.
\newblock Unsupervised salience learning for person re-identification.
\newblock In {\em IEEE CVPR}, 2013.

\bibitem{zhao2014learning}
R.~Zhao, W.~Ouyang, and X.~Wang.
\newblock Learning mid-level filters for person re-identification.
\newblock In {\em IEEE CVPR}, 2014.

\bibitem{zheng2015scalable}
L.~Zheng, L.~Shen, L.~Tian, S.~Wang, J.~Wang, J.~Bu, and Q.~Tian.
\newblock Scalable person re-identification: A benchmark.
\newblock In {\em IEEE ICCV}, 2015.

\bibitem{zheng2011person}
W.-S. Zheng, S.~Gong, and T.~Xiang.
\newblock Person re-identification by probabilistic relative distance
  comparison.
\newblock In {\em IEEE CVPR}, 2011.

\bibitem{zheng2013reidentification}
W.-S. Zheng, S.~Gong, and T.~Xiang.
\newblock Reidentification by relative distance comparison.
\newblock {\em IEEE TPAMI}, 35(3):653--668, 2013.

\bibitem{zhengtowards}
W.-S. Zheng, S.~Gong, and T.~Xiang.
\newblock Towards open-world person re-identification by one-shot group-based
  verification.
\newblock {\em IEEE TPAMI}, PP(99):1--1, 2015.

\end{thebibliography}
}
\end{document}